\def\year{2020}\relax
\documentclass[letterpaper]{article} 
\usepackage{aaai20}  
\usepackage{times}  
\usepackage{helvet} 
\usepackage{courier}  
\usepackage[hyphens]{url}  
\usepackage{graphicx} 
\usepackage{amsmath}
\usepackage{graphicx}  
\title{Question Generation by Transformers}
\author{Kettip Kriangchaivech\textsuperscript{\rm 1}\thanks{To whom correspondence should be addressed.} and Artit Wangperawong\textsuperscript{\rm 2}\\ 
\textsuperscript{\rm 1}kettipk@gmail.com \\
\textsuperscript{\rm 2}artit.wangperawong@usbank.com \\  
 U.S. Bank \\
1095 Avenue of the Americas\\
New York, NY 10036 
}
 
\begin{document}

\maketitle

\begin{abstract}
A machine learning model was developed to automatically generate questions from Wikipedia passages using transformers, an attention-based model eschewing the paradigm of existing recurrent neural networks (RNNs). The model was trained on the inverted Stanford Question Answering Dataset (SQuAD), which is a reading comprehension dataset consisting of 100,000+ questions posed by crowdworkers on a set of Wikipedia articles. After training, the question generation model is able to generate simple questions relevant to unseen passages and answers containing an average of 8 words per question. The word error rate (WER) was used as a metric to compare the similarity between SQuAD questions and the model-generated questions. Although the high average WER suggests that the questions generated differ from the original SQuAD questions, the questions generated are mostly grammatically correct and plausible in their own right.
\end{abstract}

\section{Introduction}

Existing question generating systems reported in the literature involve human-generated templates, including cloze type \cite{hermann2015teaching}, rule-based \cite{mitkov2003computer,rus2010first}, or semi-automatic questions \cite{alvaro2010linked,rey2012semi,liu2014sherlock}. On the other hand, machine learned models developed recently have used recurrent neural networks (RNNs) to perform sequence transduction, i.e. sequence-to-sequence \cite{du2017learning,kim2019improving}. In this work, we investigated an automatic question generation system based on a machine learning model that uses transformers instead of RNNs \cite{transformers,artitw}. Our goal was to generate questions without templates and with minimal human involvement using machine learning transformers that have been demonstrated to train faster and better than RNNs. Such a system would benefit educators by saving time to generate quizzes and tests.

\section{Background and Related Work}

A relatively simple method for question generation is the fill-in-the-blank approach, which is also known as cloze tasks. Such a method typically involves the sentence first being tokenized and tagged for part-of-speech with the named entity or noun part of the sentence masked out. These generated questions are an exact match to the one in the reading passage except for the missing word or phrase. Although fill-in-the-blank questions are often used for reading comprehension, answering such questions correctly may not necessarily indicate comprehension if it is too easy to match the question to the relevant sentence in the passage. To improve fill in the blank type questions, a prior study used a supervised machine learning model to generate fill-in-the-blank type questions. The model paraphrases the sentence from the passage with the missing word by anonymizing entity markers \cite{hermann2015teaching}. 

Semi-automatic methods can also be use for question generation. Semi-automatic question generation involves human-generated templates in combination with querying the linked database repositories to complete the question \cite{alvaro2010linked,rey2012semi}. The answer to the question is also extracted from the linked database. If the question is to be answered selecting from multiple choices, then distractors could also be selected from the database and randomly generated as incorrect choices for the answer. Another example of template-based question-and-answer generator using linked data is called Sherlock that has been shown to generate questions with varying levels of difficulty \cite{liu2014sherlock}. However, designing a large set of high quality questions using semi-automatic question generation methods can be cognitively demanding and time-consuming. The types of questions created are also constrained to the templates. Generating a large dataset of questions is therefore cumbersome.

Other automatic question generators require human-made rules for the model to follow \cite{mitkov2003computer,rus2010first}. Educators are recruited to define the rules that will convert declarative sentences into interrogative questions \cite{wang2007automatic,Adamson2013AutomaticallyGD,heilman-smith-2010-good}. The rules generated requires the educator to possess both linguistic knowledge and subject knowledge. As with the template-based methods described above, this rules-based method can also be time-consuming and cognitively demanding. Moreover, the quality of the questions is limited by the quality of the handcrafted rules, and rules-based approaches are not scalable beyond human capacity.

Perhaps the most automated method reported thus far utilizes RNNs as sequence transduction (seq2seq) models to generate questions from sentences or passages \cite{du2017learning,kim2019improving}. In the most successful variant of RNNs, the Long Short-Term Memory (LSTM) networks, the model reads from left to right and includes an encoder and a decoder \cite{keneshloo2018deep}. The encoder takes the input and converts it to hidden vectors, while the decoder takes the vectors from the encoder and creates its own hidden vector to predict the next word based on the previous hidden vector \cite{keneshloo2018deep}. The hidden vector to the decoder stores all of the information about the context. The components in between the encoder and the decoder of the seq2seq model consists of attention, beam search, and bucketing. The attention mechanism takes the input to the decoder and allows the decoder to analyze the input sequence selectively. Beam search mechanism allows the decoder to select the highest probability word occurrence based on previous words. The bucketing mechanism allows the length of sequences to vary based on what we designate the bucket size to be. The decoder is then rewarded for correctly predicting the next word and penalized for incorrect predictions.

In this study, we developed a seq2seq model to automatically generate questions from Wikipedia passages. Our goal is to produce plausible questions with minimal human intervention that can assist educators in developing their quizzes and tests. Our model is based on transformers instead of RNNs. Transformers can train faster than RNNs because it is more parallelizable, working well with large and limited datasets \cite{transformers,artitw}. 

Transformers can also achieve better performance at a fraction of the training cost. Like the RNN approach, transformers have an encoder and a decoder. Transformers also incorporate the beam search and bucketing mechanisms. Unlike RNNs, transformers adopt multiple attention heads without requiring any recurrence, though recurrence can be added. The self-attention mechanism used is the scaled dot-product attention according to
\begin{equation}\label{eq:1}
    Attention(K,Q,V)=softmax\left(\frac{QK^T}{\sqrt{d}}\right)V,
\end{equation}
\noindent where $d$ is the dimension (number of columns) of the input queries $Q$, keys $K$, and values $V$. By using self-attention, transformers can account for the whole sequence in its entirety and bidirectionally. For multi-head attention with $h$ heads that jointly attend to different representation subspaces at different positions given a sequence of length $m$ and the matrix $H\in\mathbf{R}^{m \times d}$, the result is

\begin{align}
\label{eq:2}
\begin{split}
 MultiHead(H) &= Concat\left(head_1,...,head_h\right)W^O ,
\\
 head_i &= Attention\left(H^W_i,H^K_i,H^V_i\right) ,
\end{split}
\end{align}

\noindent where the projections are learned parameter matrices $H^W_i,H^K_i,H^V_i\in\mathbf{R}^{(d \times d)/h}$ and $W^O\in\mathbf{R}^{(d \times d)}$. 

Models utilizing transformers have achieved state-of-the-art performance on many NLP tasks, including question answering \cite{BERT,XLNet}. It is therefore interesting to study how transformers might be used to generate questions by training on the inverted SQuAD.

\section{Experimental Methods}

\subsubsection{Data.} In this study, we used the Stanford Question Answering Dataset (SQuAD). SQuAD is a reading comprehension dataset consisting of 100,000+ questions posed by crowdworkers on a set of Wikipedia articles, where the answer to each question is a segment of text from the corresponding reading passage \cite{rajpurkar2016squad}. To generate the data for SQuAD, the top 10,000 English Wikipedia articles were ranked by Project Nayuki's Wikipedia's internal PageRanks as high-quality. Paragraphs that are longer than 500 characters were then extracted from the articles and partitioned into a training set (80\%), a development set (10\%), and a test set (10\%). Only the former two datasets are publicly available. Crowdworkers were then employed to generate the questions and then the answers to the questions based on the extracted paragraphs. Another subset of crowdworkers were then asked to answer the questions that were generated given the corresponding passage to compare the model's answer with human generated answers and provide a benchmark for machine learning models.

\subsubsection{Pre-processing.} We used the publicly available data from SQuAD to train our model to generate the questions. We used SQuAD's training and dev sets as our training and test sets, respectively. The reading passage, question, and answer data were pre-processed as described in the next section. For the test set, we provided the model with the pre-processed reading passages and answers that were never seen by the model. We inverted SQuAD by training a machine learning model to infer a question given a reading passage and an answer separated by a special token (i.e., `*') as input.

\begin{table*}[t]
\centering
\caption{Named entity tags and the entities they encompass.}\smallskip
\begin{tabular}{l|l}

    \hline
    \textbf{Base Tag} & \textbf{Description}\\ 
    \hline
    
    PERSON & People, including fictional.\\ \hline
    NORP  & Nationalities or religious or political groups.\\ \hline
    FAC & Buildings, airports, highways, bridges, etc.\\ \hline
    ORG & Companies, agencies, institutions, etc.\\ \hline
    GPE & Countries, cities, states.\\ \hline
    LOC & Non-GPE locations, mountain ranges, bodies of water.\\ \hline
    PRODUCT & Objects, vehicles, foods, etc. (Not services.)\\ \hline
    EVENT & Named hurricanes, battles, wars, sports events, etc.\\ \hline
    WORK\_OF\_ART & Titles of books, songs, etc.\\ \hline
    LAW & Named documents made into laws.\\ \hline
    LANGUAGE & Any named language.\\ \hline
    DATE & Absolute or relative dates or periods.\\ \hline
    TIME & Times smaller than a day.\\ \hline
    PERCENT & Percentage, including ``\%".\\ \hline
    MONEY & Monetary values, including unit.\\ \hline
    QUANTITY & Measurements, as of weight or distance.\\ \hline
    ORDINAL & ``first", ``second", etc.\\ \hline
    CARDINAL & Numerals that do not fall under another type.\\ \hline
    
\end{tabular}
\label{table:1}
\end{table*}

For pre-processing the reading passages, questions and answers, spaCy was used for named entity recognition and part-of-speech tagging \cite{spacy2}, and WordPiece was used for tokenization \cite{wordpiece}. To ensure intelligible outputs, stop words are removed from the context passages and answers but not the questions. After lowercasing, tokenizing and removing the stop words, the named entities are then replaced with their respective tags to better allow the model to generalize and learn patterns in the data. We address a variety of named and numeric entities, including companies, locations, organizations and products, etc. (Table~\ref{table:1}). To account for multiple occurrences of a named entity type in the context passage, we also included an index after the named entity tag. As an example, consider the following context passage: 

\begin{quote}
Super Bowl 50 was an American football game to determine the champion of the National Football League (NFL) for the 2015 season. The American Football Conference (AFC) champion Denver Broncos defeated the National Football Conference (NFC) champion Carolina Panthers 24–10 to earn their third Super Bowl title. The game was played on February 7, 2016, at Levi's Stadium in the San Francisco Bay Area at Santa Clara, California. As this was the 50th Super Bowl, the league emphasized the "golden anniversary" with various gold-themed initiatives, as well as temporarily suspending the tradition of naming each Super Bowl game with Roman numerals (under which the game would have been known as "Super Bowl L"), so that the logo could prominently feature the Arabic numerals 50.
\end{quote}

\noindent Applying the pre-processing described above, including the indexed-named entity tag replacement but not yet removing stop words, would produce

\begin{quote}
EVENT 0 DATE 0 was an NORP 0 football game to determine the champion of ORG 0 ( ORG 1 ) for DATE 1 . the NORP 0 football conference ( ORG 2 ) champion ORG 3 defeated ORG 4 ( ORG 5 ) champion ORG 6 24 – 10 to earn their ORDINAL 0 EVENT 0 title . the game was played on DATE 2 , at FAC 0 in FAC 1 at GPE 0 , GPE 1 . as this was the ORDINAL 1 EVENT 0 , the league emphasized the " golden anniversary " with various gold - themed initiatives , as well as temporarily suspend \#\#ing the tradition of naming each EVENT 0 game with LANGUAGE 0 nu \#\#meral \#\#s ( under which the game would have been known as " EVENT 0 l " ) , so that the logo could prominently feature the LANGUAGE 1 nu \#\#meral \#\#s DATE 0 .
\end{quote}

\noindent Note that the index is separated from the named entity tag by a space and therefore interpreted by the model as a separate token so that named entities of similar types can be associated and generalized from without sacrificing the ability to distinguish between different entities of the same type. This spacing is necessary since we do not employ character-level embedding. 

To generate the sub-word embeddings, we used the pre-trained WordPiece model from BERT, which has a 30,000 token vocabulary. WordPiece is a statistical technique used to segment text into tokens at the sub-word level. The vocabulary is initialized with all individual characters and iteratively aggregates the most frequently and likely combination of symbols into a vocabulary. The generation of WordPieces allows the model to capture the meaning of commonly occurring word segments, such as the root word \textit{suspend} from \textit{suspending} in the example context passage above. This dispenses with the need for the model to learn different variants or conjugations of a word. 

Each input to the model comprised of a concatenation of the pre-processed answer and context passage. The most commonly agreed upon answer was chosen from among the three plausible answers for each question in SQuAD. Unlike prior question generation studies using SQuAD by isolating the sentence containing the answer, here we include the entire passage because the answers can depend on the context outside of the answer-containing sentence. Compared to RNNs used in prior studies, transformers allow us to more conveniently train and perform inference on longer sequence lengths. The model was developed with TensorFlow \cite{tensorflow} and Tensor2Tensor \cite{tensor2tensor}, and then trained with an Nvidia Tesla T4 GPU for 1 million training steps. 

To evaluate and analyze the results, the generated questions are post-processed by removing unnecessary spaces and consolidating the resulting WordPieces into a single coherent word. In other words, the results were de-tokenized using BERT's pre-trained WordPiece model.

\section{Results and Discussion}
To measure the model's question formulating ability, we calculated the word error rate (WER) between the generated questions and the corresponding questions from SQuAD. The SQuAD questions were used as the reference questions, as they ideally ask for the answers provided. WER is also known as the edit distance, which is a measure of similarity at the word level between generated questions and the target questions from SQuAD. 

In essence, WER is the Levenshtein distance applied to a sequence of words. Differences between the two sequences can include sequence insertions, deletions and substitutions. WER can be calculated according to  

\begin{equation}\label{eq:3}
    WER=\frac{(S + D + I)}{N}=\frac{(S + D + I)}{(S + D + C)},
\end{equation}

\noindent where $S$ is the number of word substitutions, $D$ is the number of word deletions, $I$ is the number of word insertions, $C$ is the number of correct words, and $N$ is the total number of words in the reference sequence (i.e., $N = S + D + C$).

\begin{figure}[!hbt]
    \centering
    \includegraphics[width=1.0\linewidth]{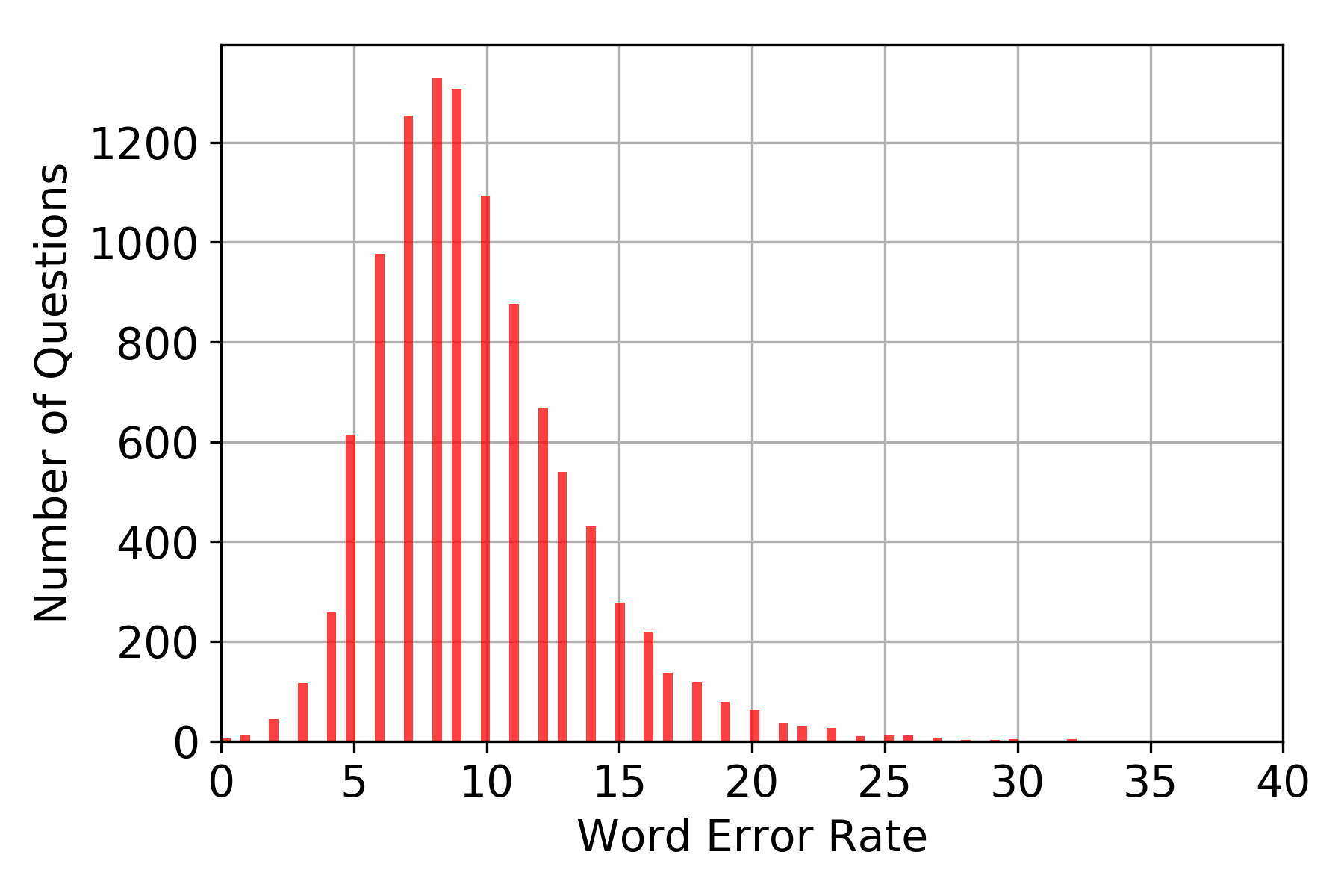}
    \caption{Word error rate (WER) between the SQuAD questions and model-generated questions on our test set (SQuAD dev set).}
    \label{fig:1}
\end{figure}

A low WER would indicate that a model-generated question is similar to the reference question, while a high WER means that the generated question differs significantly from the reference question. It should be noted that a high WER does not necessarily invalidate the generated question, as different questions can have the same answers, and there could be various ways of phrasing the same question. On the other hand, a situation with low WER of 1 could be due to the question missing a pertinent word to convey the correct idea. Despite these shortcomings in using WER as a metric of success, the WER can reflect our model's effectiveness in generating questions that are similar to those of SQuAD based on the given reading passage and answer. WER can be used for initial analyses that can lead to deeper insights as discussed further below.

\begin{figure}[!hbt]
    \centering
    \includegraphics[width=1.0\linewidth]{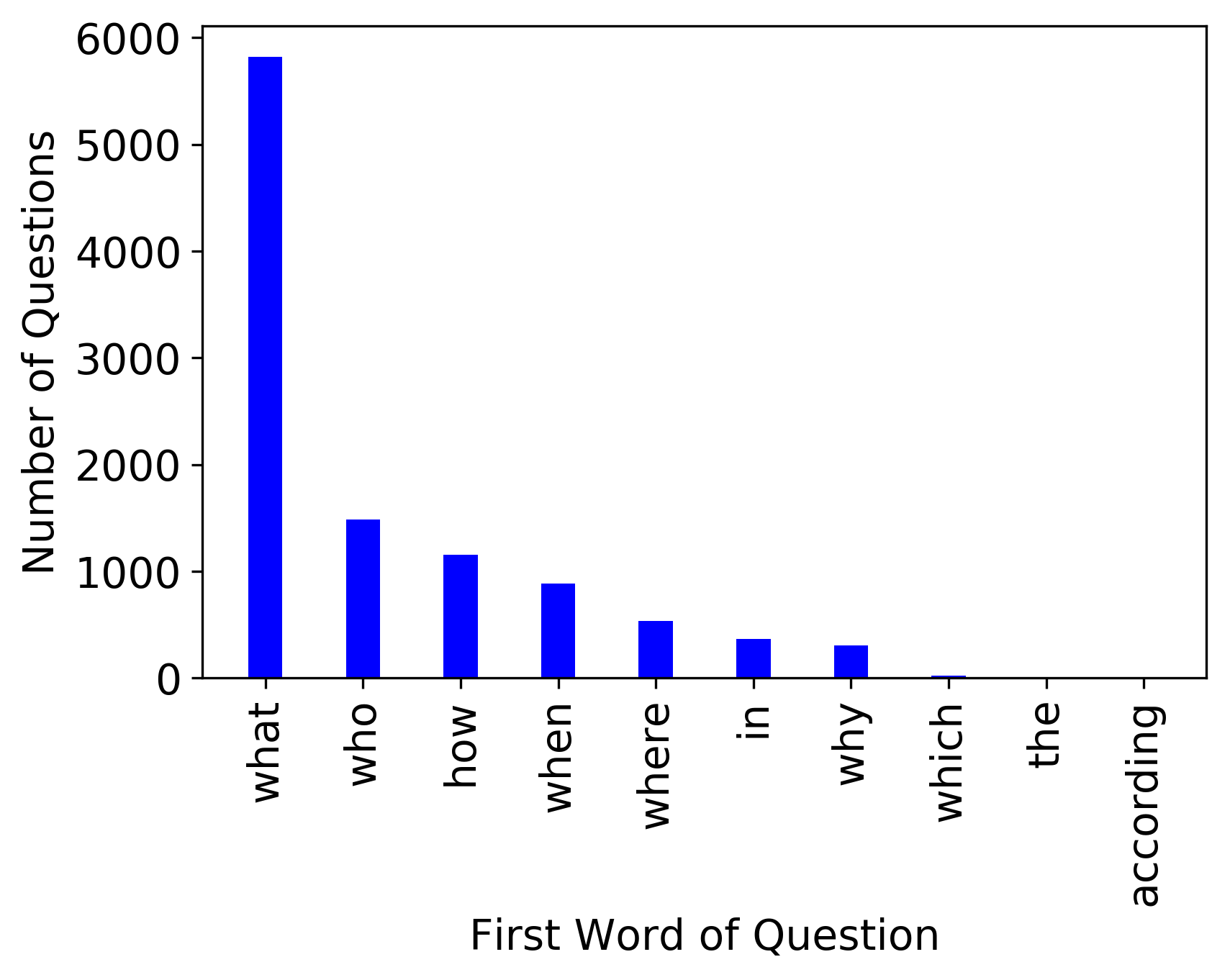}
    \caption{First word frequency of the model-generated questions in descending order for our test set (SQuAD dev set). Results for all first words are shown. }
    \label{fig:2}
\end{figure}

\begin{figure}[!hbt]
    \centering
    \includegraphics[width=1.0\linewidth]{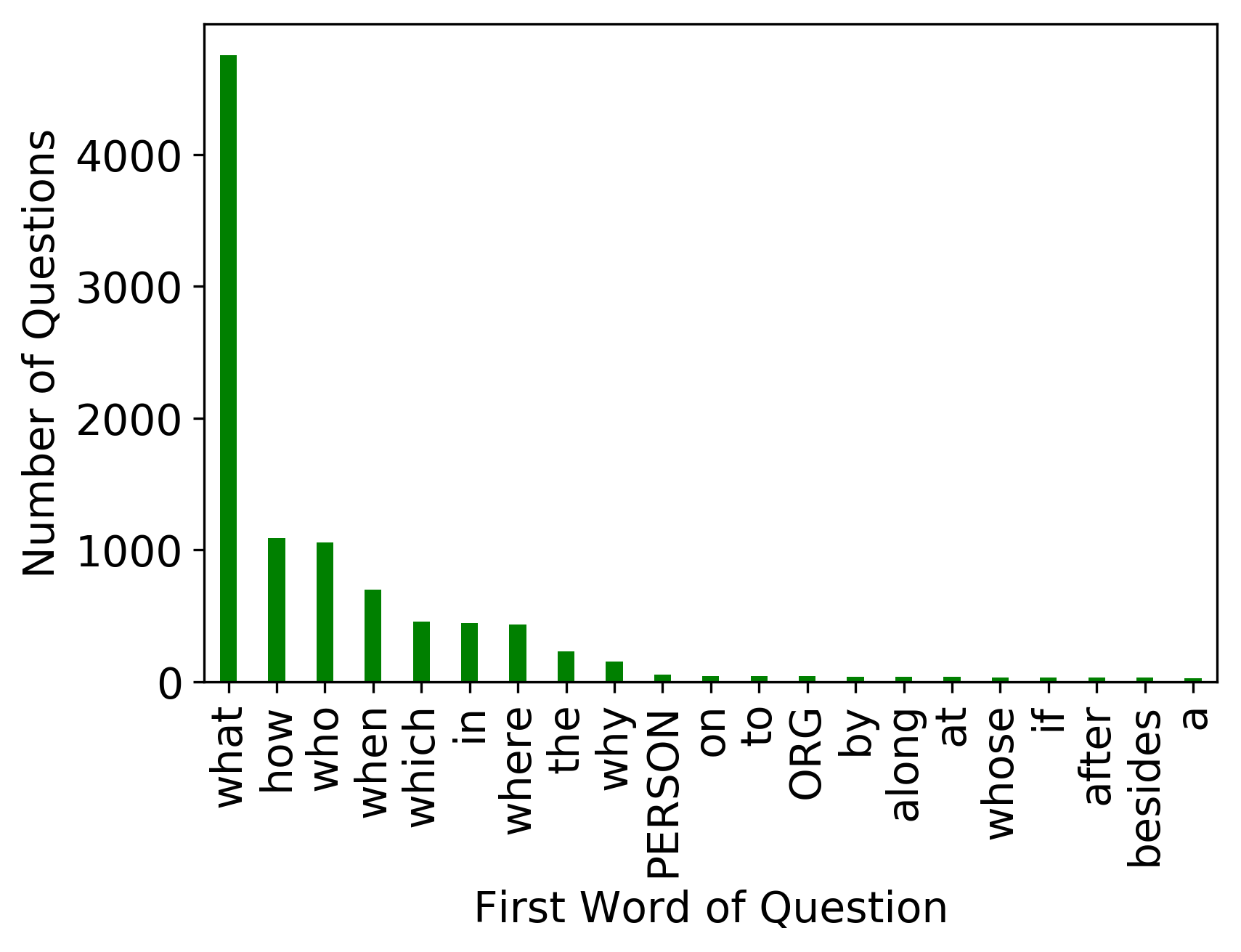}
    \caption{First word frequency of the SQuAD questions in descending order for the dev set. Only the 21 most frequent first words are shown. The first words all capitalized, i.e. \textit{PERSON} and \textit{ORG}, are named entity tags assigned after pre-processing.}
    \label{fig:3}
\end{figure}

\begin{figure}[!hbt]
    \centering
    \includegraphics[width=1.0\linewidth]{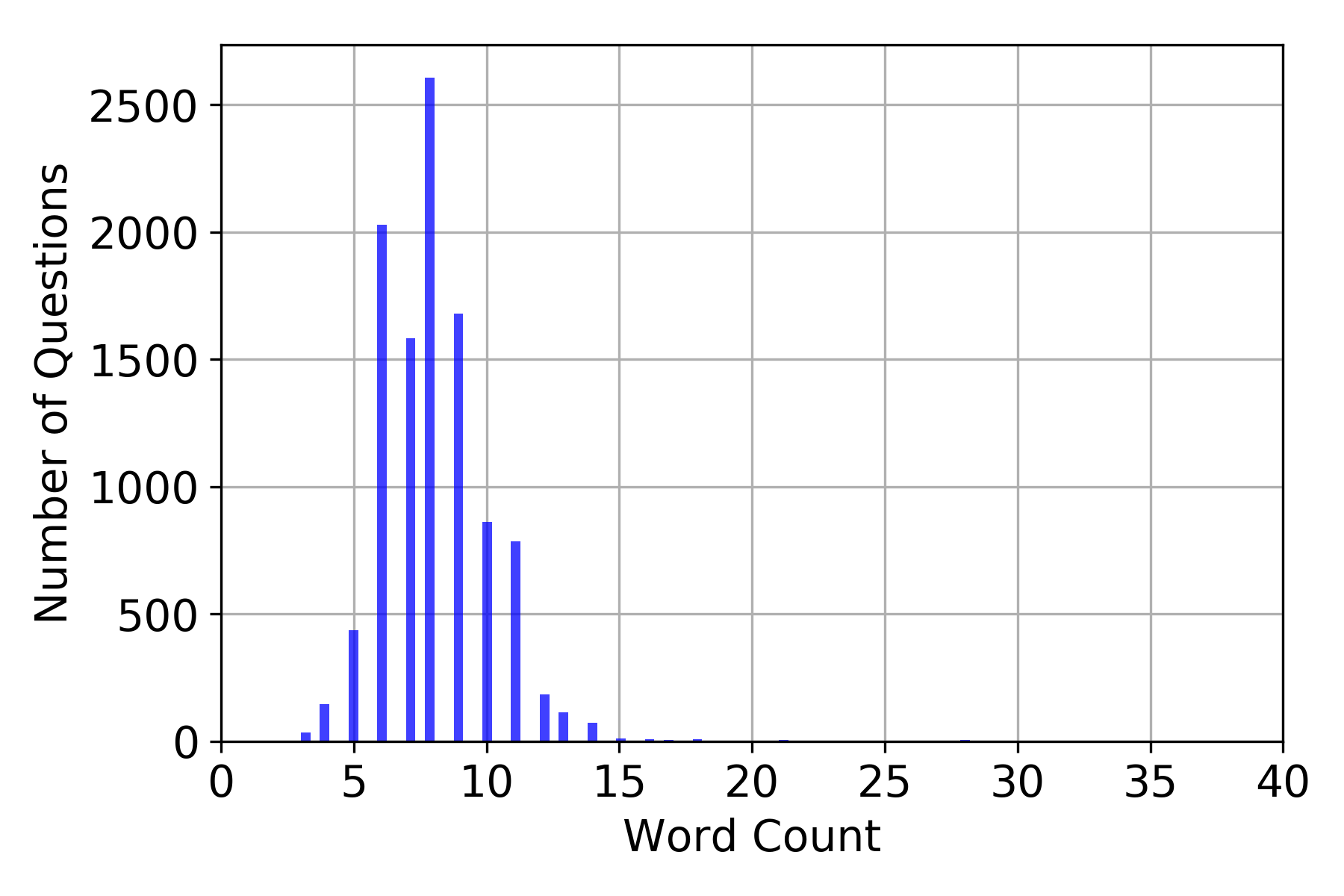}
    \caption{Word count histogram of the model-generated questions for our test set (SQuAD dev set).}
    \label{fig:4}
\end{figure}

\begin{figure}[!hbt]
    \centering
    \includegraphics[width=1.0\linewidth]{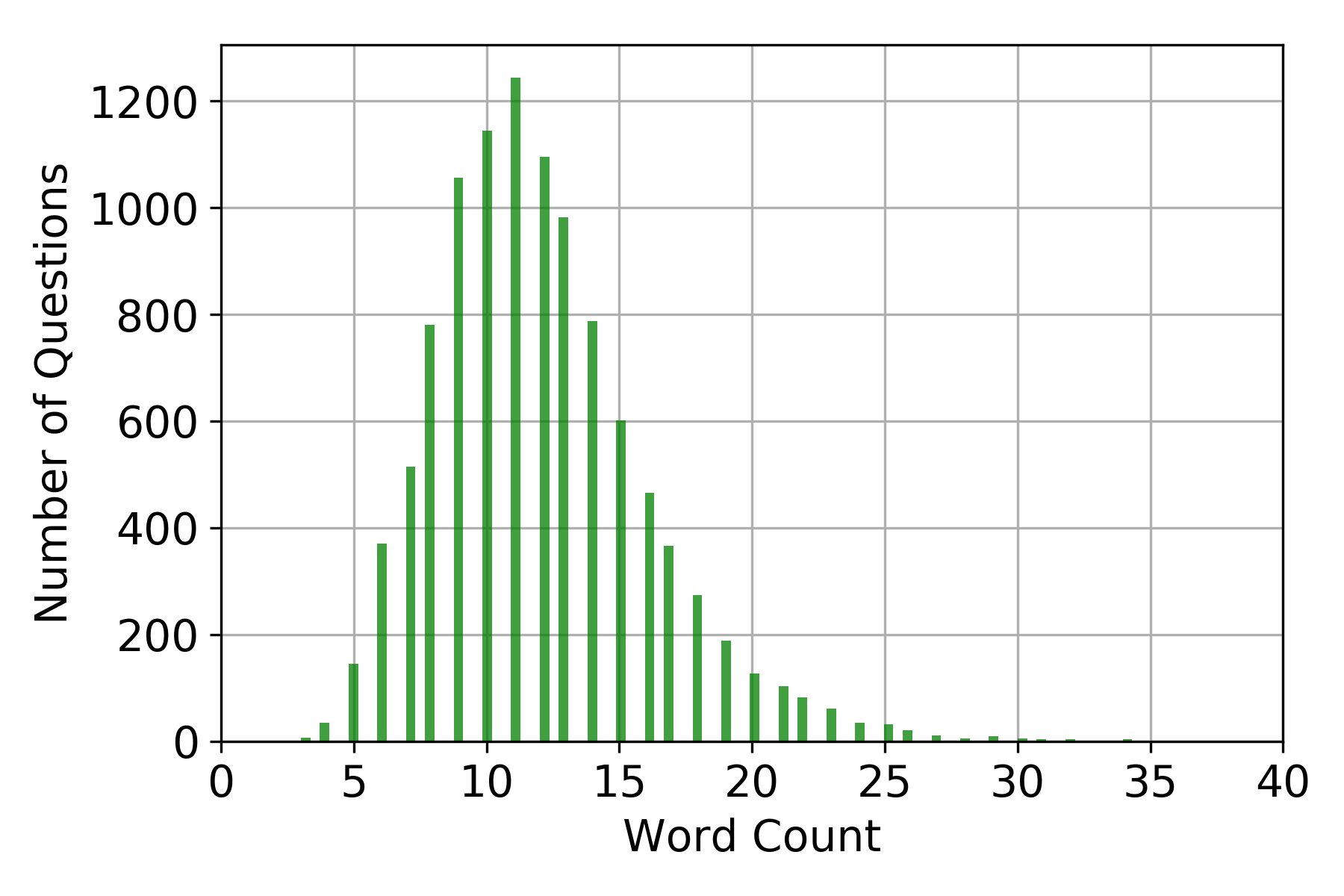}
    \caption{Word count histogram of the SQuAD questions for our test set (SQuAD dev set).}
    \label{fig:5}
\end{figure}

\begin{table*}[t]
\centering
\caption{Select SQuAD and model-generated questions with various word error rates (WERs). The model-generated questions have been de-tokenized using BERT's pre-trained WordPiece model.}\smallskip
\resizebox{1.0\textwidth}{!}{ 
\begin{tabular}{l|l|c}

    \hline
    \textbf{SQuAD} & \textbf{Model-generated} & \textbf{WER}\\ 
    \hline
    where was PERSON 8 born? & where was PERSON 8 born? & 0\\ \hline
    how many museums are in GPE 0? & how many museums are in GPE 0? & 0\\ \hline
    what is the largest city of GPE 1? & what is the largest area of GPE 1? & 1\\ \hline
    where did PERSON 2 die? & when did PERSON 2 die? & 1\\ \hline
    where did PERSON 4 die? & how did PERSON 4 die? & 1\\ \hline
    where was ORG 0 located? & where was ORG 2 located? & 1\\ \hline
    where is ORG 0 based? & where is ORG 0 located? & 1\\ \hline
    who is the president of GPE 3? & who was the president of GPE 3? & 1\\ \hline
    when did the launches of boilerplate csms occur in orbit? & when was the ORDINAL 0 satellite launched? & 8\\ \hline
    by what century did researchers see that they could liquefy air? & in what century did water begin? & 9\\ \hline
    by what means were scientists able to liquefy air? & what is the name of the process of water? & 9\\ \hline
    what other european NORP 4 leader was educated at ORG 1? & who was the leader of GPE 5? & 10\\ \hline
    what type of engines became popular for power generation after piston steam engines? & what type of electric motors have? & 10\\ \hline
    what was an example of a type of warship that required high speed? & what was the name of the NORP 0 ships? & 10\\ \hline
    
    what happens to the gdp growth of a country if the income share of the top & what is the average amount of gdp? & 22 \\ 
    PERCENT 0 increases,according to imf staff economists? & ~ & ~ \\ \hline
    
    if the average GPE 1 worker were to complete an additional year of school,& what was the average income rate of GPE 0? & 23 \\
    what amount of growth would be generated over 5 years? & ~ & ~ \\ \hline
    
\end{tabular}
}
\label{table:2}
\end{table*}

Using the SQuAD dev set as our test set, our model generated 10,570 questions. The questions generated were mostly grammatically correct and related to the topic of the context passage. Fig.~\ref{fig:1} shows the WER distribution, which has a mean of 9.66 words. 0.05\% of the total questions generated by the model were an exact match to the corresponding SQuAD questions. For our discussion and analysis, we examine generated questions with various different WERs to gain insight into the quality of the questions. 9.94\% of the model-generated questions have a WER less than or equal to 5, 56.38\% have a WER between 6 and 10, 26.41\% with a WER between 11 and 15, 5.81\% with a WER between 16 and 20, and 1.45\% of the generated questions have a WER greater than 21 (Fig.~\ref{fig:1}). In addition to the WER, the number of words in the questions (Figs.~\ref{fig:4}~and~\ref{fig:5}) and the first word of the questions (Figs.~\ref{fig:2}~and~\ref{fig:3}) are also considered below.

As shown in Fig.~\ref{fig:1}, our model was able to generate the exact questions as those corresponding to SQuAD with a WER of 0 for a small portion of instances. These types of questions tend to be relatively simpler and shorter, which are easier to learn as apparent in select examples from Table~\ref{table:1}. The model was also able to learn about and utilize synonyms, as apparent in the following examples where the WER is 1. Instead of using \textit{based} as in the target question from SQuAD, ``where is ORG 0 based?", the model used \textit{located} to generate ``where is ORG 0 located?". Although the term \textit{area} can encompass many meanings, \textit{city} and \textit{area} can be synonymous depending on the context and therefore ``what is the largest area of GPE 1?" generated by the model has the same meaning as ``what is the largest city of GPE 1?" from SQuAD. The ability to capture relative meaning between words indicates that applying a pre-trained contextualized language model can improve performance in future studies.

Beyond exact matches, a low WER does not guarantee that a model-generated question has the same meaning as the target SQuAD question. Consider the following examples where the WERs are all 1, but the meanings differ between the generated and target questions. Sometimes the model produced a question in past tense when the target question from SQuAD is in present tense, e.g. ``who was the president of GPE 3?" generated by the model versus ``who is the president of GPE 3?" from SQuAD. In some cases, the named entity type matched, but the index did not, e.g. ``where was ORG 2 located?" generated by the model versus ``where was ORG 0 located?" from SQuAD. This case of swapping the index of a given named entity type could be a consequence of the pre-processing employed. Despite the different meanings, the questions generated are plausible questions that could be reasonably asked given the same context passages.

Since the questions generated by our model on the SQuAD dev set have an average WER of 9.66, we examined the questions with a WER of 8 to 10 (Table~\ref{table:2}) to see whether the questions are structured properly and whether they maintain the same meaning as the target SQuAD questions. As shown in Table~\ref{table:2}, it was challenging for the model to produce questions as complex as those from SQuAD, which resulted in a large WER. The average word count for the generated questions is 8 words per question (Fig.~\ref{fig:4}), while most of the questions from SQuAD are more detailed and complex with a total average word count of 12 (Fig.~\ref{fig:5}). The structure of the questions generated by the model are simpler and less detailed than those from SQuAD, but most are nevertheless plausible questions that are grammatically correct. The model-generated questions are relevant to the topic of the context passage and has the correct type of asking words to start the question. For example, although the target question from SQuAD is ``by what means were scientists able to liquefy air?", the model can generate ``what is the name of the process of water?". The questions have a WER of 9 and both can be interpreted as referring to the transformation process of water; more specifically, the condensation process. The model may have not sufficiently learned about the term \textit{liquefy} but was still able to ask a similar question given limitations in the dataset used for training. 

We next examined questions in the high WER regime, i.e. WER of 20 or more. Questions generated by the model still reflect the answer and context passage of interest. For example, given inputs for the target SQuAD question ``who was responsible for the authorship of a paper published on real time - computations?", the model generated ``what did PERSON 3 write?". Although both questions ask about authorship, the model's question is asking for a different type of answer, as indicated by the asking word of the questions (i.e., \textit{who} vs. \textit{what}). 

To understand how the model chooses the asking word, we plot the first-word frequencies in descending order for SQuAD and model-generated questions (Figs.~\ref{fig:2}~and~\ref{fig:3}). Questions from SQuAD predominantly involve \textit{what}, which reflects the first-word distribution of the training set as well. As the first-word is usually the asking word, training data imbalance most likely caused the model to be biased towards generating \textit{what} questions. While the questions from SQuAD involve over 21 different words that initiate the questions (Figs.~\ref{fig:3}), our model only uses 10 different words to initiate questions (Figs.~\ref{fig:2}). The lack of diversity demonstrates that our model is not as well versed in asking questions as the crowdworkers in SQuAD and forms less elaborate types of questions. After all, our model generates an average of 8 words per question (Fig.~\ref{fig:4}), whereas the SQuAD questions have an average of 12 words per question (Fig.~\ref{fig:5}).

\section{Conclusion and Future Work}

We demonstrate that a transformer model can be trained to generate questions with correct grammar and relevancy to the context passage and answers provided. WER analyses was applied to diagnose shortcomings and guide future improvements. We observed that a low WER could be due to syntactic similarity but semantic disagreement, while two questions with syntactic divergence but similar meaning could result in a high WER. Since our results does not exhibit issues relating to contextual and syntactic roles of words within a generated question, other popular metrics (BLEU, ROUGE, F1-score, etc.) would lead to similar findings  \cite{wer_he}. Perhaps a better approach to evaluating question generation models is to apply state-of-the-art question answering models from SQuAD's leaderboard to measure how many answers agree.

To improve the model, more and balanced data can be provided to train the model to reduce the asking word bias. One method that can be used to obtain more data is through data augmentation by back-translation \cite{dab}. The original SQuAD can be translated into another language such as French. The translated text could then be translated back into English to generate a variation of the context, question and answers that provide more training data for the model. Another data augmentation method is to paraphrase SQuAD's data to get another variation of the text \cite{summarization}, but one would have to ensure that pertinent information is not sacrificed in the summarization. The augmented data should then be sampled to reduce bias as much as possible. Other pre-processing variations can be considered. We tried including stopwords and removing the answer form the context passages but did not see improvements. Recent advancements in pre-trained bidirectionally contextualized language models can also be incorporated \cite{BERT,XLNet}, which would require a decoder to be added to the pre-trained model.

\bibliographystyle{aaai}
\bibliography{citation.bib}

\begin{thebibliography}{}

\bibitem[\protect\citeauthoryear{Abadi \bgroup et al\mbox.\egroup
  }{2015}]{tensorflow}
Abadi, M.; Agarwal, A.; Barham, P.; Brevdo, E.; Chen, Z.; Citro, C.; Corrado,
  G.~S.; Davis, A.; Dean, J.; Devin, M.; Ghemawat, S.; Goodfellow, I.; Harp,
  A.; Irving, G.; Isard, M.; Jia, Y.; Jozefowicz, R.; Kaiser, L.; Kudlur, M.;
  Levenberg, J.; Man\'{e}, D.; Monga, R.; Moore, S.; Murray, D.; Olah, C.;
  Schuster, M.; Shlens, J.; Steiner, B.; Sutskever, I.; Talwar, K.; Tucker, P.;
  Vanhoucke, V.; Vasudevan, V.; Vi\'{e}gas, F.; Vinyals, O.; Warden, P.;
  Wattenberg, M.; Wicke, M.; Yu, Y.; and Zheng, X.
\newblock 2015.
\newblock {TensorFlow}: Large-scale machine learning on heterogeneous systems.
\newblock Software available from tensorflow.org.

\bibitem[\protect\citeauthoryear{Adamson \bgroup et al\mbox.\egroup
  }{2013}]{Adamson2013AutomaticallyGD}
Adamson, D.; Bhartiya, D.; Gujral, B.; Kedia, R.; Singh, A.; and Ros{\'e},
  C.~P.
\newblock 2013.
\newblock Automatically generating discussion questions.
\newblock In {\em AIED}.

\bibitem[\protect\citeauthoryear{{\'A}lvaro and
  {\'A}lvaro}{2010}]{alvaro2010linked}
{\'A}lvaro, G., and {\'A}lvaro, J.
\newblock 2010.
\newblock A linked data movie quiz: the answers are out there, and so are the
  questions.

\bibitem[\protect\citeauthoryear{Devlin \bgroup et al\mbox.\egroup
  }{2018}]{BERT}
Devlin, J.; Chang, M.; Lee, K.; and Toutanova, K.
\newblock 2018.
\newblock {BERT:} pre-training of deep bidirectional transformers for language
  understanding.
\newblock {\em CoRR} abs/1810.04805.

\bibitem[\protect\citeauthoryear{Du, Shao, and Cardie}{2017}]{du2017learning}
Du, X.; Shao, J.; and Cardie, C.
\newblock 2017.
\newblock Learning to ask: Neural question generation for reading
  comprehension.
\newblock {\em arXiv preprint arXiv:1705.00106}.

\bibitem[\protect\citeauthoryear{{He}, {Deng}, and {Acero}}{2011}]{wer_he}
{He}, X.; {Deng}, L.; and {Acero}, A.
\newblock 2011.
\newblock Why word error rate is not a good metric for speech recognizer
  training for the speech translation task?
\newblock In {\em 2011 IEEE International Conference on Acoustics, Speech and
  Signal Processing (ICASSP)},  5632--5635.

\bibitem[\protect\citeauthoryear{Heilman and
  Smith}{2010}]{heilman-smith-2010-good}
Heilman, M., and Smith, N.~A.
\newblock 2010.
\newblock Good question! statistical ranking for question generation.
\newblock  609--617.
\newblock Los Angeles, California: Association for Computational Linguistics.

\bibitem[\protect\citeauthoryear{Hermann \bgroup et al\mbox.\egroup
  }{2015}]{hermann2015teaching}
Hermann, K.~M.; Kocisky, T.; Grefenstette, E.; Espeholt, L.; Kay, W.; Suleyman,
  M.; and Blunsom, P.
\newblock 2015.
\newblock Teaching machines to read and comprehend.
\newblock In {\em Advances in neural information processing systems},
  1693--1701.

\bibitem[\protect\citeauthoryear{Honnibal and Montani}{2017}]{spacy2}
Honnibal, M., and Montani, I.
\newblock 2017.
\newblock spacy 2: Natural language understanding with bloom embeddings,
  convolutional neural networks and incremental parsing.
\newblock \url{https://spacy.io}.

\bibitem[\protect\citeauthoryear{Keneshloo \bgroup et al\mbox.\egroup
  }{2018}]{keneshloo2018deep}
Keneshloo, Y.; Shi, T.; Ramakrishnan, N.; and Reddy, C.~K.
\newblock 2018.
\newblock Deep reinforcement learning for sequence to sequence models.
\newblock {\em arXiv preprint arXiv:1805.09461}.

\bibitem[\protect\citeauthoryear{Kim \bgroup et al\mbox.\egroup
  }{2019}]{kim2019improving}
Kim, Y.; Lee, H.; Shin, J.; and Jung, K.
\newblock 2019.
\newblock Improving neural question generation using answer separation.
\newblock In {\em Proceedings of the AAAI Conference on Artificial
  Intelligence}, volume~33,  6602--6609.

\bibitem[\protect\citeauthoryear{Liu and Lin}{2014}]{liu2014sherlock}
Liu, D., and Lin, C.
\newblock 2014.
\newblock Sherlock: a semi-automatic quiz generation system using linked data.
\newblock In {\em International Semantic Web Conference (Posters \& Demos)},
  9--12.
\newblock Citeseer.

\bibitem[\protect\citeauthoryear{Mitkov and Ha}{2003}]{mitkov2003computer}
Mitkov, R., and Ha, L.~A.
\newblock 2003.
\newblock Computer-aided generation of multiple-choice tests.
\newblock In {\em Proceedings of the HLT-NAACL 03 workshop on Building
  educational applications using natural language processing},  17--22.

\bibitem[\protect\citeauthoryear{Rajpurkar \bgroup et al\mbox.\egroup
  }{2016}]{rajpurkar2016squad}
Rajpurkar, P.; Zhang, J.; Lopyrev, K.; and Liang, P.
\newblock 2016.
\newblock Squad: 100,000+ questions for machine comprehension of text.
\newblock {\em arXiv preprint arXiv:1606.05250}.

\bibitem[\protect\citeauthoryear{Rey \bgroup et al\mbox.\egroup
  }{2012}]{rey2012semi}
Rey, G.~{\'A}.; Celino, I.; Alexopoulos, P.; Damljanovic, D.; Damova, M.; Li,
  N.; and Devedzic, V.
\newblock 2012.
\newblock Semi-automatic generation of quizzes and learning artifacts from
  linked data.

\bibitem[\protect\citeauthoryear{Rus \bgroup et al\mbox.\egroup
  }{2010}]{rus2010first}
Rus, V.; Wyse, B.; Piwek, P.; Lintean, M.; Stoyanchev, S.; and Moldovan, C.
\newblock 2010.
\newblock The first question generation shared task evaluation challenge.

\bibitem[\protect\citeauthoryear{See, Liu, and Manning}{2017}]{summarization}
See, A.; Liu, P.~J.; and Manning, C.~D.
\newblock 2017.
\newblock Get to the point: Summarization with pointer-generator networks.
\newblock {\em arXiv preprint arXiv:1704.04368}.

\bibitem[\protect\citeauthoryear{Vaswani \bgroup et al\mbox.\egroup
  }{2017}]{transformers}
Vaswani, A.; Shazeer, N.; Parmar, N.; Uszkoreit, J.; Jones, L.; Gomez, A.~N.;
  Kaiser, L.; and Polosukhin, I.
\newblock 2017.
\newblock Attention is all you need.
\newblock {\em CoRR} abs/1706.03762.

\bibitem[\protect\citeauthoryear{Vaswani \bgroup et al\mbox.\egroup
  }{2018}]{tensor2tensor}
Vaswani, A.; Bengio, S.; Brevdo, E.; Chollet, F.; Gomez, A.~N.; Gouws, S.;
  Jones, L.; Kaiser, L.; Kalchbrenner, N.; Parmar, N.; Sepassi, R.; Shazeer,
  N.; and Uszkoreit, J.
\newblock 2018.
\newblock Tensor2tensor for neural machine translation.
\newblock {\em CoRR} abs/1803.07416.

\bibitem[\protect\citeauthoryear{Wang, Hao, and Liu}{2007}]{wang2007automatic}
Wang, W.; Hao, T.; and Liu, W.
\newblock 2007.
\newblock Automatic question generation for learning evaluation in medicine.
\newblock In {\em International conference on web-based learning},  242--251.
\newblock Springer.

\bibitem[\protect\citeauthoryear{Wangperawong}{2018}]{artitw}
Wangperawong, A.
\newblock 2018.
\newblock Attending to mathematical language with transformers.
\newblock {\em CoRR} abs/1812.02825.

\bibitem[\protect\citeauthoryear{Wu \bgroup et al\mbox.\egroup
  }{2016}]{wordpiece}
Wu, Y.; Schuster, M.; Chen, Z.; Le, Q.~V.; Norouzi, M.; Macherey, W.; Krikun,
  M.; Cao, Y.; Gao, Q.; Macherey, K.; Klingner, J.; Shah, A.; Johnson, M.; Liu,
  X.; Kaiser, L.; Gouws, S.; Kato, Y.; Kudo, T.; Kazawa, H.; Stevens, K.;
  Kurian, G.; Patil, N.; Wang, W.; Young, C.; Smith, J.; Riesa, J.; Rudnick,
  A.; Vinyals, O.; Corrado, G.; Hughes, M.; and Dean, J.
\newblock 2016.
\newblock Google's neural machine translation system: Bridging the gap between
  human and machine translation.
\newblock {\em CoRR} abs/1609.08144.

\bibitem[\protect\citeauthoryear{Xie \bgroup et al\mbox.\egroup }{2019}]{dab}
Xie, Q.; Dai, Z.; Hovy, E.~H.; Luong, M.; and Le, Q.~V.
\newblock 2019.
\newblock Unsupervised data augmentation.
\newblock {\em CoRR} abs/1904.12848.

\bibitem[\protect\citeauthoryear{Yang \bgroup et al\mbox.\egroup
  }{2019}]{XLNet}
Yang, Z.; Dai, Z.; Yang, Y.; Carbonell, J.~G.; Salakhutdinov, R.; and Le, Q.~V.
\newblock 2019.
\newblock Xlnet: Generalized autoregressive pretraining for language
  understanding.
\newblock {\em CoRR} abs/1906.08237.

\end{thebibliography}

\end{document}